\documentclass{article}

\usepackage{arxiv}

\usepackage[utf8]{inputenc} % allow utf-8 input
\usepackage[T1]{fontenc}    % use 8-bit T1 fonts
\usepackage{hyperref}       % hyperlinks
\usepackage{url}            % simple URL typesetting
\usepackage{booktabs}       % professional-quality tables
\usepackage{amsfonts}       % blackboard math symbols
\usepackage{nicefrac}       % compact symbols for 1/2, etc.
\usepackage{microtype}      % microtypography
\usepackage{lipsum}		% Can be removed after putting your text content
\usepackage{graphicx}
\usepackage{natbib}
\usepackage{doi}
\usepackage{ulem}
\usepackage{amsmath}
\usepackage{listings}
 \usepackage{verbatim} 

\usepackage{color}
\definecolor{lbcolor}{rgb}{0.9,0.9,0.9}  
\title{Toychain: A Simple Blockchain for Research in Swarm Robotics}

\author{ 
    \href{https://orcid.org/0000-0000-0000-0000}{\includegraphics[scale=0.06]{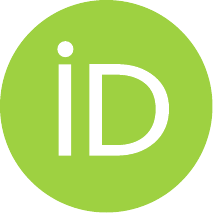}\hspace{1mm}Alexandre Pacheco}\thanks{alexandre.melo.pacheco@gmail.com}\\
    Universit\'{e} Libre Bruxelles\\
    \And
	Ulysse Denis\\
    Universit\'{e} Libre Bruxelles\\
    \And
	\href{https://orcid.org/0000-0002-2419-3789}{\includegraphics[scale=0.06]{orcid.pdf}\hspace{1mm}Raina Zakir}\\
    Universit\'{e} Libre Bruxelles\\
	\AND
	\href{https://orcid.org/0000-0003-2974-9827}{\includegraphics[scale=0.06]{orcid.pdf}\hspace{1mm}Volker Strobel}\\
    Universit\'{e} Libre Bruxelles\\
	\And
	\href{https://orcid.org/0000-0003-4745-992X}{\includegraphics[scale=0.06]{orcid.pdf}\hspace{1mm}Andreagiovanni Reina}\\
	University of Konstanz\\
	\And
	\href{https://orcid.org/0000-0002-3971-0507}{\includegraphics[scale=0.06]{orcid.pdf}\hspace{1mm}Marco Dorigo}\\
	Universit\'{e} Libre Bruxelles\\
}

\renewcommand{\headeright}{Technical Report}
\renewcommand{\undertitle}{Technical Report}
\renewcommand{\shorttitle}{Toychain}

\hypersetup{
pdftitle={Toychain: A Simple Blockchain for Research in Swarm Robotics},
pdfsubject={},
pdfauthor={Alexandre Pacheco, Ulysse Denis, Volker Strobel, Andreagiovanni Reina, Marco Dorigo},
pdfkeywords={},
}

\begin{document}
\maketitle

\begin{abstract}
\noindent
This technical report describes the implementation of Toychain: a simple, lightweight blockchain implemented in Python, designed for ease of deployment and practicality in robotics research. It can be integrated with various software and simulation tools used in robotics (we have integrated it with ARGoS, Gazebo, and ROS2), and also be deployed on real robots capable of Wi-Fi communications. The Toychain package supports the deployment of smart contracts written in Python (computer programs that can be executed by and synchronized across a distributed network). The nodes in the blockchain can execute smart contract functions by broadcasting transactions, which update the state of the blockchain upon agreement by all other nodes. The conditions for this agreement are established by a consensus protocol. The Toychain package allows for custom implementations of the consensus protocol, which can be useful for research or meeting specific application requirements. Currently, Proof-of-Work and Proof-of-Authority are implemented.
\end{abstract}

% \keywords{First keyword \and Second keyword \and More}

\section{Introduction}

In today's digital world, ensuring data security and integrity is paramount. Blockchain technology has emerged as a revolutionary solution to this concern by enabling fully decentralized networks to agree on the state of information stored in the blockchain. At their core, blockchains are simple data structures that record ordered transactions in cryptographically linked data blocks~\citep{Nak2008:techreport}. A consensus protocol enables large-scale networks to agree on the same version of the blockchain, establishing a global state from the sequential execution of transactions in its blocks. Smart contracts, which are computer programs stored on the blockchain, can host a user-defined state machine. This extends the applicability of blockchain technology beyond Bitcoin's original deployment as a financial ledger, particularly, showing great promise for applications in multi-robot systems~\citep{DorPacReiStr2024:natrevee}.

Researchers have been exploring blockchain applications for swarm robotics~\citep{StrCasDor2018:aamas,StrCasDor2020:frontiers,PacStrDor2020:ants,PacStrReiDor2022:ants,StrPacDor2023:sciencerobotics,ZhaPacStr-etal2023:iros,VanPacStr-etal2023:SciRep}, but existing blockchain protocols often fail to account for the dynamic network topologies and limited computational capabilities of robots. Moreover, these blockchain protocols may not integrate seamlessly with common simulation software. Most blockchain protocols are designed for global-scale internet networks, potentially making research-oriented simulations with hundreds of nodes cumbersome. Additionally, modifying existing blockchain repositories to meet research requirements is often time-consuming and undesirable (for example, to implement custom consensus protocols, or to speed up simulations by using a simulation clock rather than the system clock). %Moreover, existing blockchain protocols do not allow for speeding up their execution, therefore, collecting data becomes time-consuming, as simulations have to be executed in real-time.

To address these challenges, we developed a simple blockchain called Toychain, tailored for research applications. Toychain simplifies aspects of blockchain technology not relevant to research, making it suitable for use in different simulators and deployable on physical robots. This eliminates the need to switch to a `real' blockchain, such as Ethereum~\citep{But2014:techreport}, when transitioning from simulation studies to physical deployments. Toychain allows for synchronizing time steps in simulators to time steps of Toychain; this way, even if the simulator is executed faster than real-time, the blockchain mechanisms (e.g., generation of blocks) are proportionally sped up and research data can thus be collected much faster than real-time. Conversely, when executed on real robots, Toychain may use the internal clocks of the robots (as long as the consensus protocol can handle asynchronous communications).

This technical report describes the complete implementation of Toychain as a Python module, detailing its installation, usage, and fine-tuning---particularly the consensus protocol, which is implemented as a plug-in module.

\section{Installation}

The GitHub repository of Toychain can be found online: \url{https://github.com/teksander/toychain}. The repository can be cloned as follows:
\begin{lstlisting}[language=bash]
    git clone https://github.com/teksander/toychain
\end{lstlisting}

A full example of how the module can be integrated with the ARGoS simulator \citep{PinTriOGr-etal2012:si} is also online \url{https://github.com/teksander/toychain-argos}, and can be cloned using:

\begin{lstlisting}[language=bash]
    git clone https://github.com/teksander/toychain-argos --recurse-submodules
\end{lstlisting}

In this repository, the Toychain is a submodule which executes automatically when running ARGoS simulations (a blockchain node is created for each robot in the simulation). This requires \hyperlink{https://github.com/ilpincy/argos3}{ARGoS} and \hyperlink{https://github.com/KenN7/argos-python}{ARGoS-python}, which can be installed following the online documentation.

A video tutorial of how to install the Toychain and write a first smart contract is also available: \url{https://www.youtube.com/watch?v=3YvWE1HWBMw}.

\section{Structure}
The following section describes the code structure of the Toychain. 

\subsection{Node}
Each node of the network is represented by the class \texttt{Node}. This class represents an agent (e.g., a human or a robot), in the network.

\subsubsection*{Chain and Mempool}
Each node has its own \texttt{chain} attribute, which corresponds to its local blockchain, and its \texttt{mempool} attribute, which represents the pending transactions it has received and are not yet included on its blockchain. The \texttt{chain} is a list of \verb|Block| objects (Section~\ref{block}), while the \texttt{mempool} is a dictionary that maps transaction unique ids to \verb|Transaction| objects (Section~\ref{transaction}).

\subsubsection*{Enode}
\label{enode}
To be able to interact with other nodes, each \texttt{Node} object has a host, a port, and a unique id. These are stored in a string, called \texttt{enode}, that serves as a node identifier in the network.
\begin{equation}
    \texttt{enode://<id>@<host>:<port>}
\end{equation}

For example, when adding a new peer on the blockchain, the input value for the \verb|add_peer()| method is the \texttt{enode}. 

\subsubsection*{Custom Timer}
Each node possesses an attribute called \verb|custom_timer| that stores its internal clock. If the Toychain is deployed on a real-time application (such as physical robots), then the \verb|custom_timer| should be set to be the Python module \texttt{time}. Otherwise, to synchronize the Toychain with the time of a simulator, the \verb|custom_timer| should be an instance of the \verb|CustomTimer| class, and needs to be be incremented at each simulation step using its \texttt{step} method. The \verb|CustomTimer| has three methods:

\begin{itemize}
    \item \verb|time()|  : returns the current time
    \item \verb|sleep(duration)| : idles the node for a period of time
    \item \verb|step()|  : increases the counter by one
\end{itemize}
\underline{Remark:} Since each node has its own \verb|CustomTimer| instance, each node must execute the step function each time the simulation clock increases.

\subsubsection*{Consensus}
The \texttt{consensus} (Section~\ref{consensus}) attribute is required when initializing the node, and it refers to the decision process by which the node will accept or reject changes to the blockchain state. It establishes the rules for the valid production of new blocks and their validation by other nodes. In Section~\ref{consensus} we describe how consensus protocols can be created and employed as plug-in modules.

\subsubsection*{Threads}
Each \texttt{Node} has three threads, which can be executed synchronously (in simulations) by executing their corresponding \texttt{step} methods once per simulation step. In a deployment on real robots, it is advisable to use the asynchronous implementation that uses Python's \texttt{threading} package. The \verb|mining_thread| is responsible for block production, and is programmed within the \texttt{consensus} class. To control the \verb|mining_thread|, the methods \verb|start_mining()| and \verb|stop_mining()| are used. 

The two other threads are called \verb|mempool_sync_thread| and \verb|chain_sync_thread|. They respectively handle the synchronization of the \texttt{mempool} and \texttt{chain} between the node and its current peers. Theses two threads automatically start when the node starts the TCP connection and stop when the TCP stops (\verb|start_tcp()| and \verb|stop_tcp()|, see Section~\ref{tcp}).

\subsection{Block}
\label{block}

The \texttt{Block} class has the following attributes : 

\begin{itemize}
    \item \verb|height|: height of this block in the chain
    \item \verb|parent_hash|: the hash of the previous block in the chain
    \item \verb|data|: list of \verb|Transaction| objects
    \item \verb|timestamp|: timestamp of the production time of the block
    \item \verb|miner_id|:  the id of the producer of the block
    \item \verb|difficulty|: the difficulty of the block used for consensus
    \item \verb|total_difficulty|: the cumulative difficulty of all blocks in the chain
    \item \verb|transactions_root|: the hash (or Merkle root) of the transactions list
    \item \verb|state|\footnote{It is not typical to record the state in the block, as each node typically maintains its own state machine. However, in a Toychain, it can be useful to store the state at each block as this can save on the computational costs of having each node in a simulator execute state transitions themselves (particularly in simulations, where computation costs are more valuable than bandwidth costs)}: optional, \hyperref[state]{\texttt{State}} object that results from applying the transactions to the state of the previous block
    \item \verb|nonce|: optional, depending on the consensus protocol
    \item \verb|hash|: the hash of the block which is computed using the \verb|compute_block_hash()| method
\end{itemize}

The \verb|data| attribute stores transactions as an ordered list, and is typically called the block body (which is the bulk of its size in terms of storage costs). Blocks additionally contain other attributes which form the block header. These attributes allow the nodes to establish whether a block meets the rules of the consensus protocol without incurring the costs of downloading and processing the block body. For example, the header contains the \verb|timestamp| of the block, and the \verb|parent_hash| which links that block to the previous block in the chain. The \verb|total_difficulty| parameter is used by the consensus mechanisms to establish the ``strongest'' chain (see \hyperref[consensus]{consensus} in Section~\ref{consensus}), which is accepted by nodes as the current version of the chain (potentially discarding other alternatives).

\subsubsection*{Transaction}
\label{transaction}
The \verb|Transaction| class has the following attributes:

\begin{itemize}
    \item \verb|sender|: the address of the sender
    \item \verb|receiver|: the address of the receiver
    \item \verb|value|: the integer value of the transaction
    \item \verb|data|: a text string (such as JSON)
    \item \verb|timestamp|: the timestamp of transaction broadcast
    \item \verb|id|:  an unique identifier for the transaction
    \item \verb|nonce|: a counter that enables senders to order their transactions 
\end{itemize}

In the \verb|data| attribute, the sender can write anything it desires to record in the blockchain, and it can also be used to encode interactions with the smart contract (see \hyperref[state]{State}). In order to be processed by the smart contract, the sender should send a JSON encoded dictionary with the keys \verb|"function"| and \verb|"inputs"|, which correspond to the smart contract function to be executed and the inputs it requires (as a list).

\subsubsection*{State}
\label{state}
Each block has a \verb|State| object that handles the global state variables. The state variables can be modified to be what is required by the consensus protocol or application. Currently, we have a state variable called \verb|balances|, which is a dictionary mapping the identities/addresses of the nodes to their balances of Toycoins (the cryptocurrency of Toychain). Users can send Toycoins to each other by sending transactions with a non-empty \verb|value| field. Additionally, we have another state variable called \verb|contract|, which stores all the state variables specific to the smart contract, which are user-programmable.

The initial \verb|State| is hardcoded in the genesis block and is the starting state for all nodes. Afterwards, the nodes update and synchronize the state by executing the transactions which are added to blocks in the blockchain.
% Each time a block is produced, the block state should be first initialised with the previous block state variables. This can be done by putting the optional input variable \verb|state_var| in the block object constructor. 
%Then, for each transaction that the block contains, 
To apply a transaction, nodes use the method \verb|state.apply_transaction()|. Like its name indicates, the method applies the specified transaction to the state variables: first it transfers the Toycoins from the sender to the receiver (provided the sender has enough of them), and then executes the instructions encoded in the \verb|data| field which will execute functions in the smart contract and therefore update the \verb|contract| state variable. 

% An additional action of \verb|apply_transaction()| is the fact that it can apply custom functions to make some kind of smart contract. In this implementation, one custom method example is shown. It is named \verb|add_k()|. It adds the value k to a state variable named \verb|n|. The way of making this method apply to the state is to put the name of the custom function as a string in the \verb|transaction.data| under the key \verb|"action"| and the input under the key \verb|"input"|. For example, putting this dictionary in a transaction data will add 4 to the state variable n : \verb|{"action": "add_k", "input": 4}|. 

% The way of defining a custom method is to define it as a method of the \verb|State| class.

\subsection{Communications}
\label{tcp}
In order to synchronize the blocks and transactions, the nodes need to communicate. The Toychain uses the Transmission Control Protocol (TCP) protocol, as it can be used on the localhost in simulations, but also extended to a local network to allow communications between physical robots. It uses the \verb|socket| library of Python. 
Each node works like a server that answers to requests made to it. To be able to synchronize, the nodes request to one another information about their blocks and their mempool at regular intervals. 

Each node has a \verb|NodeServerThread|. This thread handles the TCP communication with other nodes. It has a defined \verb|(host, port)| and listens to connection requests. When it receives a connection request, it reads the request and sends back a message with the requested information. To handle the message syntax and content, each \verb|NodeServerThread| has a \verb|MessageHandler|. The \verb|MessageHandler| returns the message to send through the TCP socket.

\subsubsection*{Peering}
The nodes only send requests to nodes that they consider as their peers. In the case of robots, we typically allow for dynamic networks, in which the robots add peers that are in close range, and remove them afterwards. To do so, nodes can add peers using the \verb|Node.add_peer()| method. Afterwards they can remove each other from their peer list with \verb|Node.remove_peer()|. The input value to both these methods is the enode of the peer (see \hyperref[enode]{enode}). 

\subsubsection*{Synchronization}
At each regular interval, each node requests information about the blockchain and mempool of its peers. This is done by the pingers. Each node has one pinger for the block synchronization and one pinger for the mempool synchronization. These pingers are threads represented by the classes \verb|MempoolPinger| and \verb|ChainPinger|. The time interval at which the nodes send requests is defined in the \verb|constants.py| file. At each interval, they start iterating on each peer to send them a request.

The mempool synchronization is the simpler one of the two: each time a request for mempool synchronization is received by a node, it answers with a list of all the transactions contained in its mempool. The receiver then adds novel transactions to its own mempool.

The block synchronization works as follows with Node~1 requesting information about blockchain of Node~2:
\begin{enumerate}
    \item Node~1 sends a request to Node~2. 
    \item Node~2 sends \verb|hash| and the  \verb|total_difficulty| of the last block in its blockchain. This communication is low cost, as it contains no actual blocks or data.
    \item Node~1 checks if its last block hash corresponds to the one specified in the answer message:
    \begin{itemize}
        \item If it does: the two chains are synchronized and no action is needed. The \verb|ChainPinger| proceeds to the next peer. 
        \item If it does not, but the total difficulty received is equal to the total difficulty of its own chain: the two chains have the same total difficulty so Node~1 keeps its chain.
        \item If it does not and the total difficulty received is lower than the total difficulty of its own chain: Node~1 will send a block request to Node~2 (see below).
    \end{itemize}
\end{enumerate}

The block request is launched when a Node knows that one of his peers has a ``better'' chain (one with higher total difficulty). The goal of this request is to find a common block between the two nodes. If we continue the example above in the case where the last block from Node~2 was not in the blockchain of Node~1 and the total difficulty of Node 2 was higher than Node~1, the block request is executed as follows:%
\begin{enumerate}
    \item Node 1 sends a block request containing the hashes of the last 5 blocks in its chain to avoid overloading the TCP connection.
    \item Node 2 iterates its chain (from the end) to find a common block:
    \begin{itemize}
        \item If it finds a common block: an answer with a partial chain (from the common block to the end) is sent.
        \item If a common block is not found: a \verb|None| answer is sent.
    \end{itemize}
    \item Node 1 looks at the answer:
    \begin{itemize}
        \item If it is a partial chain: it is appended to its chain starting from the common block (see below)
        \item If the answer is None: another request is sent, but with the 5 blocks further in its chain.
    \end{itemize}
\end{enumerate}
\underline{Remark:} All these procedures are handled in the \verb|MessageHandler| class.

When a node appends a partial chain to its blockchain, it first checks again that the total difficulty of the received partial chain is lower than its own (it could happen that during the process of request, a new block was produced that increased the total difficulty). Then, it checks that the chain received respects the consensus rules. This verification depends on the consensus used and can be customised. \textbf{If the received chain does not respect the consensus rules, it is dropped. }
After being sure that the partial chain is correct, it removes all the transactions from its mempool that are present in the partial chain. 
When the common block does not correspond to the last block of Node 1, it means that a chain fork occurred and that Node 1 should drop all blocks after the common block (in favor of the received partial chain). In this case, all the transactions present in the blocks are placed back in the mempool, if they are not already present in the received partial chain.

Only after these steps, the partial chain is finally added to Node 1's blockchain. These steps correspond to the method \verb|Node.sync_chain()|. 

% \subsubsection*{Message syntax}
% Each message sent through the network needs to be a dictionary that has the following key values:
% \begin{itemize}
%     \item data: It is the content of the message.  
%     \item type: It is the type of message. It determines the way the message is handled when received. There are 3 different types: the mempool sync, the chain sync and the block request. The different type strings are contained in the \verb|constant.py|.
%     \item receiver/sender of the message. Can be useful for debugging and knowing the nodes interactions. 
% \end{itemize}

% \subsubsection*{Pickle}
% Because the socket library requires to send data in a bytes object format, the pickle library is used to convert the data into a correct format using \verb|pickle.dumps()| and \verb|pickle.loads()| (see \href{https://docs.python.org/3/library/pickle.html}{the pickle documentation}). The pickle library is not able to convert class objects but is able to convert other data structures, like lists and dictionaries. To avoid this problem when sending objects like transactions or blocks, translation function have been implemented : \verb|block_to_list()|, \verb|create_block_from_list()|, \verb|transaction_to_dict()|, \verb|dict_to_transaction()|. They transform the initial object into a list/dictionary so that it can be transformed into pickle data afterwards. 

\section{Consensus}
\label{consensus}

When initializing a \texttt{Node} object, an argument \texttt{consensus} is required, which defines the rules for block production and verification, and provides the starting point for the blockchain. This is what enables the nodes to propagate and synchronize their blockchains successfully.

The \texttt{consensus} protocol is represented by a class that must have at least the following elements:
\begin{itemize}
    \item \verb|genesis|: genesis block that will be the first block of every node using this consensus protocol.
    \item \verb|block_generation|: is a process that will produce blocks according to the consensus rules.
    \item \verb|verify_chain(chain, previous_state)|: a method to verify whether a (partial) chain follows the rules.
\end{itemize}

For the time being, two consensus protocols are implemented: \textit{Proof-of-Work} and \textit{Proof-of-Authority}.

\subsection{Proof-of-Work}
Proof-of-Work (PoW) is a consensus protocol that is widely used in blockchain technology. It involves spending significant amounts of computational power in order to find a solution to a puzzle. The network participants, known as miners, are incentivized to spend this power because they may be rewarded with tokens of a cryptocurrency. The solved problem serves as a proof that the miner has expended real-world resources (in the form of electricity and computing power) to validate the block and add it to the blockchain. To be accepted in the blockchain, a block has to respect a certain difficulty target. The higher the difficulty target, the greater the average computing time needed to produce a block. 

\subsubsection*{Genesis}

The genesis block must include the following initial parameters:
\begin{itemize}
    \item \verb|MINING_DIFFICULTY|: represents the difficulty of mining and is thus related to the time taken to produce a block. 
    \item \verb|ProofOfWork.trust|: a flag that reduces computational overhead by letting nodes trust the state computed by other nodes (allowing them to copy the \texttt{state} recorded in the blocks, rather than performing the transactions themselves). 
\end{itemize}

\subsubsection*{Block production}
The block production of PoW is called mining. When mining, the node changes the nonce of the block until the block hash meets the difficulty target.

\begin{lstlisting}[language=bash]
    %% PoW block mining as pseudo-code
    
    while (block.hash > target_difficulty):
        block.nonce++
    end while
    
    chain.append(block)
\end{lstlisting}

% \begin{figure}[h!]
% \begin{verbatim}
%     while (block.hash > target_difficulty):
%         block.nonce++
%     endwhile
    
%     chain.append(block)
% \end{verbatim}
% \caption{POW block mining as pseudo-code}
% \end{figure}

% In this project, the target difficulty is a 256 bit-long string. The difficulty determines the number of zeroes at the beginning of the binary target. So for example, if the difficulty is 5, the target variable is:
% \begin{equation}
%     \text{target\_difficulty(5)} = \underbrace{000001111...1}_{\text{256 bits}}
% \end{equation}
% This means that the difficulty of each block is the same and thus that the criterion to prefer one chain to another is equivalent to choosing the longest chain. 

At each iteration, the block is updated: the block data is updated with the latest mempool, and the block timestamp is set to the current time. The state is regenerated to accommodate the transactions in the block. These updates are done by the \verb|update_block| method. 

% The thread that handles the block production is called the \verb|MiningThread|. When it starts, it starts mining and appends blocks to the chain once they meet the required difficulty. Once a block has been added, the loop just restarts with the next block until this new block meets the required difficulty, etc. 

\subsubsection*{Chain Verification}
The chain verification occurs every time a node checks if a received partial chain respects the consensus used. For PoW, the method \verb|verify_chain()| iterates over the blocks of the chain with the following steps:
\begin{enumerate}
    \item It checks that the parent hash is the same as the previous block hash in the chain.
    \item It checks that the timestamp is higher than the previous block. 
    \item It checks the difficulty level.
    \item It checks the block, via the method \verb|verify_block()|. It first checks that the block state is coherent with the transactions that it contains. Then, it checks that the difficulty of the consensus is respected. 
\end{enumerate}
To be able to verify the first block of the partial chain, the method \verb|verify_chain()| requires the block that precedes the first block in the partial chain.

\subsection{Proof-of-Authority}
Proof-of-Authority (PoA) is a low-cost permissioned consensus protocol that follows the description \hyperlink{https://eips.ethereum.org/EIPS/eip-225}{\textit{EIP-225: Clique proof-of-authority consensus protocol}}. The basic idea of PoA is that only certain nodes---called authorized signers---can produce blocks. Unlike in PoW, the block producers do not need to compete or spend computational resources. Another specificity of PoA is that there is a minimum time difference between two consecutive blocks in a chain called the \verb|BLOCK_PERIOD|.

In PoA, signers are responsible for producing a block during each ``round,'' with each round being separated by a specific time interval known as \verb|BLOCK_PERIOD|. At each round, a preferred signer is designated in a round-robin fashion. A distinguishing factor between blocks produced by preferred signers (in-turn) and those produced by non-preferred signers (no-turn) is the block difficulty.

The block difficulty is a metric used to differentiate between in-turn and no-turn blocks. Specifically, a block produced in-turn by the preferred signer has a higher difficulty compared to a block produced by a non-preferred signer. In the current project implementation, the difficulty for in-turn blocks (\verb|DIFF_INTURN|) is set to 2, while the difficulty for no-turn blocks (\verb|DIFF_NOTURN|) is set to 1. It is worth noting that these difficulty values can be adjusted or modified according to the specific requirements of the project.

\subsubsection*{Genesis}
The genesis block must include the following initial parameters:
\begin{itemize}
    \item \verb|BLOCK_PERIOD|: Minimum difference between two consecutive block’s timestamps.
    \item \verb|DIFF_INTURN|: Block difficulty for blocks containing in-turn signatures (by the preferred producer of this turn)
    \item \verb|DIFF_NOTURN|:  Block difficulty for blocks containing no-turn signatures (not by the preferred producer of this turn)
    \verb|DELAY_NOTURN|
    \item \verb|ProofOfAuth.trust|: Determines if the state should be checked or not when verifying a chain
\end{itemize}

\subsubsection*{Block Production}
To produce a block, every authorized signer sleeps the time defined by the \verb|BLOCK_PERIOD|. If the signer is out of turn, it waits additional time \verb|DELAY_NOTURN|. This small strategy will ensure that the in-turn signer has a slight advantage to sign and propagate versus the out-of-turn signers. 

After having waited the right amount of time, the signer produces a block by adding all the content of his mempool in the block and by putting the difficulty corresponding to his turn. Once the block is produced and added to the chain, the signer goes back to sleep.

\subsubsection*{Chain Verification}
The chain verification occurs every time a node wants to check if a certain chain respects the consensus used. For PoW, the method \verb|verify_chain()| iterates on the blocks of the chain with the following steps:
\begin{enumerate}
    \item It checks that the parent hash is the same as the previous block hash in the chain.
    \item It checks the timestamp difference between two consecutive blocks
    \item It checks the difficulty level
    \item It checks the block, via the method \verb|verify_block()|. It first checks that the block state is coherent with the transactions that it contains. Then, it checks that the difficulty of the consensus is respected. 
\end{enumerate}
As in PoW, to be able to verify the first block of the partial chain, the method \verb|verify_chain| needs the block preceding the first block of the partial chain.

\subsection{Consensus Plug-ins}
One of the main properties of the Toychain is the ability to allow for the creation of custom consensus protocols in Python. The following section will explain in detail the procedure to create a consensus that will work with the rest of the module.

The consensus created should be a class. When a \verb|Node| is created, the consensus object is given to the construction method of the \verb|Node|. 

The following attributes are mandatory:
\begin{itemize}
    \item \verb|verify_chain(chain, previous_state)|: Method that verifies that the specified chain respects the consensus. It is called each time a node is going to append/insert a partial chain is his. The \verb|chain| is a list of \verb|Block| objects. The \verb|previous_state| is the state of the previous block. It is a \verb|State| object. The method should return a boolean value (\verb|True| if the chain is correct, \verb|False| if not).
    \item \verb|genesis|: Genesis block that will be the first block of every node using this consensus. It should be a \verb|Block| object. It is recommended to put the height to zero. 
    \item \verb|block_generation|: This is a thread object that will produce blocks respecting the consensus protocol. In real robots, the block generation should be a class that inherits from the \verb|threading.Thread| class. In simulations, a different class with a \verb|step()| method is used. It should also have \verb|start()| and \verb|stop()| methods. To add a block to the chain, executing \verb|node.chain.append(block)| is sufficient after generating a block (the other synchronization processes take over from that point). 
\end{itemize}

\section{Future Work}

%As Toychain is still in the early development stage, it will need certain adaptations and improvements to meet research requirements. 
We invite others to contribute to the code repository \url{https://github.com/teksander/toychain}.

One area of improvement is to integrate more consensus algorithms. Tendermint~\citep{buchman2018tendermint}, Raft~\citep{raft}, and practical Byzantine fault tolerance~\citep{Castro1999} are interesting alternatives to Proof-of-Authority. It may also be important to have the ability to support multiple smart contracts, each with its own state.

While we have successfully integrated Toychain with ARGoS, Gazebo, and ROS2 (e.g., in \citep{MorPacStr-etal2024:ants}), it would be beneficial to see it used with other other simulators and robots used by the robotics community. Although Toychain can theoretically be used on real robots, this has yet to be attempted. We invite researchers to use Toychain with their robots and compare the results with their simulation studies to analyze the reality gap.

\section*{Acknowledgments}

Volker Strobel and Marco Dorigo acknowledge support from the Belgian F.R.S.-FNRS. A.R. acknowledges support from DFG under Germany's Excellence Strategy - EXC 2117 - 422037984.

\bibliographystyle{unsrtnat}+
\bibliography{references}  %%% Uncomment this line and comment out the ``thebibliography'' section below to use the external .bib file (using bibtex) .

\end{document}